\documentclass[sigconf,preprint]{acmart}

\usepackage{multirow}
\usepackage{graphicx}

\AtBeginDocument{%
  }

\setcopyright{acmlicensed}
\copyrightyear{2026}
\acmYear{2026}

\begin{document}

\title{Neural Adjoint Method for Meta-optics: Accelerating Volumetric Inverse Design via Fourier Neural Operators}

\author{Chanik Kang}
\authornote{Both authors contributed equally to this research.}
\affiliation{%
  \institution{Hanyang University, Department of Artificial Intelligence}
  \city{Seoul}
  \country{Korea}}
\email{chanik@hanyang.ac.kr}

\author{Hyewon Suk}
\authornotemark[1]
\affiliation{%
  \institution{Hanyang University, Department of Artificial Intelligence Semiconductor Engineering}
  \city{Seoul}
  \country{Korea}}
\email{hyewon01@hanyang.ac.kr}

\author{Haejun Chung}
\authornote{Corresponding author.}
\affiliation{%
  \institution{Hanyang University, Department of Electronic Engineering}
  \city{Seoul}
  \country{Korea}}
\email{haejun@hanyang.ac.kr}

\renewcommand{\shortauthors}{Kang et al.}

\begin{abstract}

Meta-optics promises compact, high-performance imaging and color routing. However, designing high-performance structures is a high-dimensional optimization problem: mapping a desired optical output back to a physical 3D structure requires solving computationally expensive Maxwell's equations iteratively. Even with adjoint optimization, broadband design can require thousands of Maxwell solves, making industrial-scale optimization slow and costly. To overcome this challenge, we propose the Neural Adjoint Method, a solver-supervised surrogate that predicts 3D adjoint gradient fields from a voxelized permittivity volume using a Fourier Neural Operator (FNO). By learning the dense, per-voxel sensitivity field that drives gradient-based updates, our method can replace per-iteration adjoint solves with fast predictions, greatly reducing the computational cost of full-wave simulations required during iterative refinement. To better preserve sensitivity peaks, we introduce a stage-wise FNO that progressively refines residual errors with increasing emphasis on higher-frequency components. We curate a meta-optics dataset from paired forward/adjoint FDTD simulations and evaluate it across three tasks: spectral sorting (color routers), achromatic focusing (metalenses), and waveguide mode conversion. Our method reduces design time from hours to seconds. These results suggest a practical route toward fast, large-scale volumetric meta-optical design enabled by AI-accelerated scientific computing.

\end{abstract}


\received{19 April 2026}

\maketitle

\section{Introduction}

\begin{figure}[htp]
    \centering
    \includegraphics[width=\linewidth]{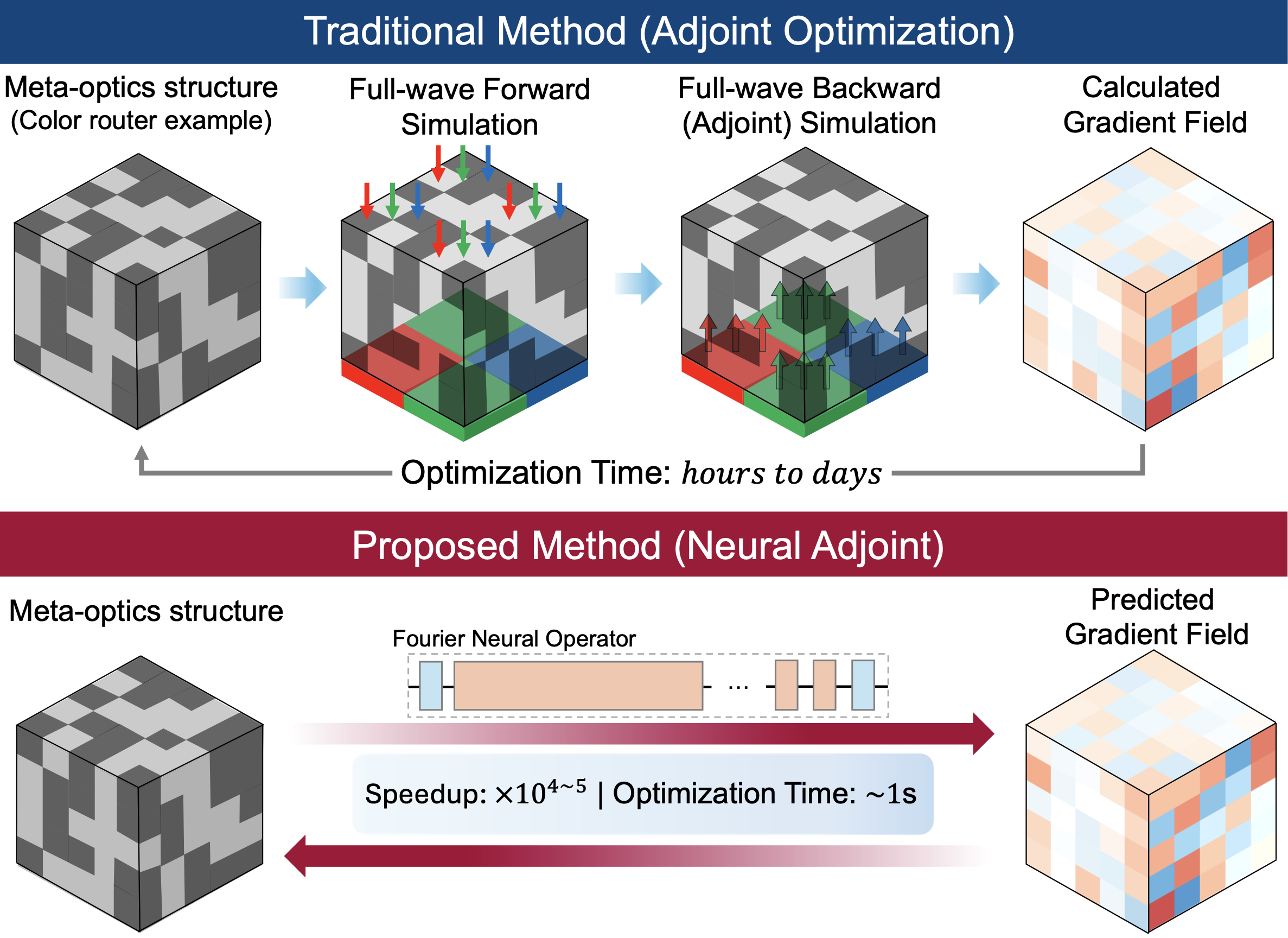}
    \caption{Breaking the simulation bottleneck in 3D meta-optics. (Top) Traditional optimization relies on expensive iterative full-wave high-fidelity simulations (Maxwell solvers) to compute gradients, taking hours or days. (Bottom) The proposed Neural Adjoint Method replaces the solver with a Fourier Neural Operator. By inferring volumetric gradients in seconds.
    }
    \label{fig:fig1}
\end{figure}

Optical-system miniaturization has accelerated the commercialization of meta-optics, indicating a major transition from academic research to industrial-scale manufacturing~\cite{kuznetsov2024roadmap, jung2026all, pestourie2018inverse}. In particular, the display and semiconductor industries are increasingly adopting 3D volumetric meta-optics, with color routers for Complementary Metal-Oxide-Semiconductor (CMOS) image sensors being a prominent example, to address the fundamental efficiency constraints of conventional dye-based color filters~\cite{lee2024inverse, park2024towards, zou2022pixel}. Unlike thin diffractive components, volumetric meta-optical devices leverage complex multiple scattering within micrometer-scale volumes to enable precise control of wavefronts and spectra. For deployment in consumer electronics, these devices must meet stringent performance requirements over broad wavelength ranges, necessitating an unprecedented degree of precision in the design and optimization process~\cite{gu2023reconfigurable}.

However, industrial deployment of these devices remains limited by substantial computational overhead. The inverse design of 3D meta-optics is typically a high-dimensional, non-convex optimization problem constrained by Maxwell’s equations~\cite{chen2022high}. The adjoint method is the standard approach for computing gradients, requiring two full-wave simulations per iteration (one forward and one adjoint)~\cite{Miller:EECS-2012-115}. Although adjoint optimization is computationally efficient in terms of gradient evaluation, the overall cost of rigorous 3D solvers such as finite-difference time-domain (FDTD) grows rapidly with problem size~\cite{kang2024large}. In addition, commercial designs must be optimized over the full visible band, which typically requires separate simulations for each wavelength or color channel~\cite{wang2021high}. As a result, a single optimization run can involve tens of thousands of full-wave simulations and take days to weeks to complete. This turnaround time poses a major obstacle for high-throughput industrial R\&D, where rapid prototyping and iterative refinement are critical.

To mitigate this cost, Deep Learning has emerged as a promising alternative. Generative models (GANs~\cite{jiang2019free, so2019designing, an2021multifunctional}, Diffusion~\cite{zhang2023diffusion, lin2025diffusion, wang2024diffmat, mao2025metamaterial}) offer rapid "one-shot" designs, but they suffer from a fundamental precision-speed trade-off~\cite{kang2024adjoint}. While fast, they often lack the physical rigor required for engineering-grade optics, producing plausible-looking structures that fail to meet the strict efficiency and fabrication tolerances of mass production. Conversely, numerical optimization guarantees physical optimality but is constrained by the extreme slowness of the solver~\cite{choi2024realization}.

To overcome this bottleneck and narrow the gap between research and manufacturing, we propose the \textbf{Neural Adjoint Method}, a real-time surrogate framework that combines the speed of generative AI with the precision of iterative optimization. Our key insight is to decouple the ``Optimizer'' from the ``Solver.'' Instead of replacing the entire optimization process with an end-to-end generator, we replace only the expensive Maxwell solver with a Fourier Neural Operator (FNO)~\cite{li2020fourier, duruisseaux2025fourier} based surrogate. This allows us to predict the 3D volumetric gradient field in a second, enabling a solver-free optimization loop that drastically shortens the design turnaround time.

Our approach addresses the challenges of volumetric inverse design through three key contributions:
\begin{itemize}
    \item \textit{Volumetric neural-adjoint gradient learning for Meta-optics.}
    We formulate inverse design around dense 3D adjoint sensitivities and train a neural operator to predict full volumetric adjoint gradient fields from voxelized devices.

    \item \textit{Stage-wise Fourier Neural Operator (SW-FNO) for multi-scale sensitivities.}
    To address the low-frequency bias of single-pass operator networks, we introduce a coarse-to-fine residual refinement scheme with progressively higher-frequency emphasis, improving recovery of localized, high-magnitude sensitivity peaks that are critical for topology refinement.

    \item \textit{Reduced-solver optimization for fast design iteration.}
    We integrate the inferred gradients into a gradient-based optimization loop, enabling rapid design iteration, and validate the approach on meta-optics problems.

\end{itemize}

We validate the proposed framework on three representative 3D inverse-design tasks: RGB color routing, achromatic metalens design, and waveguide mode conversion. Across all tasks, the Neural Adjoint Method achieves performance comparable to rigorous FDTD-based adjoint optimization, while reducing end-to-end design time by several orders of magnitude, with speedups of up to $\sim 10^{4}$--$10^{5}\times$. By mitigating the bottleneck of repeated full-wave simulations, these results suggest a practical pathway toward rapid, large-scale 3D meta-optical design enabled by AI-accelerated scientific computing.

\section{Problem Formulation}

In this section, we formulate the volumetric inverse design problem as a constrained optimization task governed by Partial Differential Equations (PDEs). We outline the computational complexities inherent in this process and introduce our Neural Adjoint formulation.


\subsection{Inverse Design for Diverse Physical Objectives}
\begin{figure*}[htp]
    \centering
    \includegraphics[width=\linewidth]{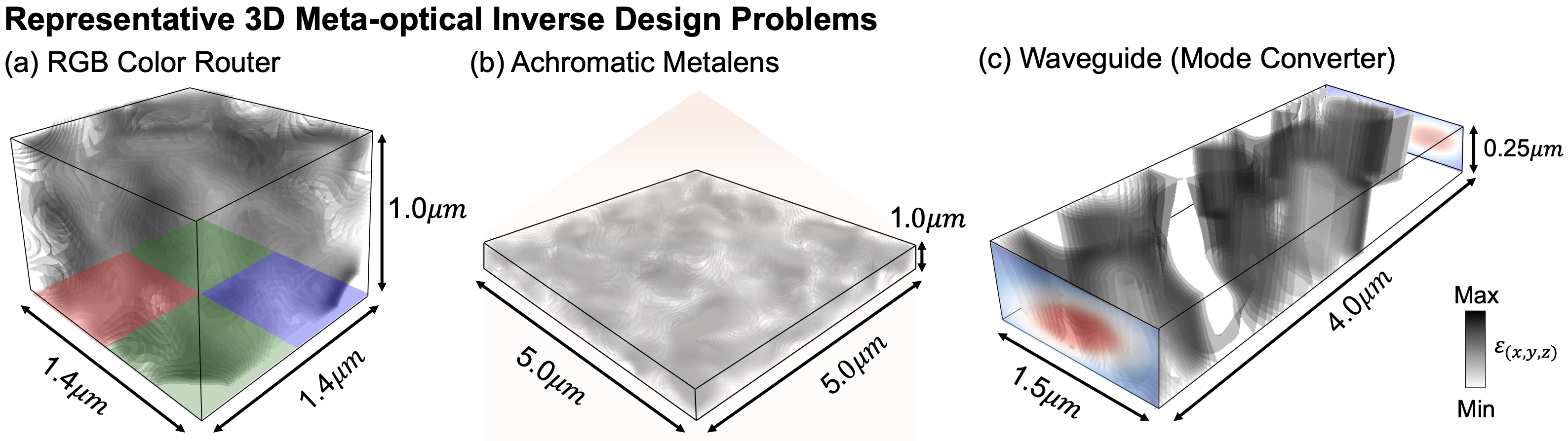}
    \caption{3D meta-optical inverse-design tasks considered in this work: (a) spectral sorting (color router), (b) light focusing (metalens), and (c) guided-wave mode conversion (waveguide).
    }
    \label{fig:fig3}
\end{figure*}

Inverse design in meta-optics and nanophotonics seeks a device parameterization (e.g., a 2D/3D permittivity distribution) that maximizes or minimizes a physics-based objective function~\cite{kang2024large}. A common approach is gradient-based adjoint optimization: given an objective functional $\mathcal{J}(\mathbf{x})$ evaluated by solving Maxwell's equations, the adjoint method computes $\nabla_{\mathbf{x}}\mathcal{J}$ using only two full-wave simulations, a forward solve to obtain the electromagnetic fields and an adjoint solve to backpropagate sensitivity, enabling scalable updates even when $\mathbf{x}$ is high-dimensional~\cite{Miller:EECS-2012-115}. In this work, we use this conventional formulation as the reference pipeline and focus on accelerating the expensive field/gradient evaluations.

As illustrated in Fig.~\ref{fig:fig1}, we consider volumetric inverse design within a three-dimensional physical domain $\Omega \subset \mathbb{R}^3$. The design variable is the spatial permittivity distribution $\epsilon(\mathbf{r})$, which is discretized as a voxel grid $\mathbf{x}\in\mathbb{R}^{N}$ with $N = N_x N_y N_z$ degrees of freedom. For a given design $\mathbf{x}$ and wavelength $\lambda$, a full-wave Maxwell solver computes the complex electric field $\mathbf{E}(\mathbf{r}; \mathbf{x}, \lambda)$, from which task-specific FoMs are defined.

We study three representative classes of meta-optics/nanophotonics objectives, which are described in Fig.~\ref{fig:fig3}:
\begin{itemize}
    \item \textbf{Intensity maximization (metalens).} We maximize the field intensity within a prescribed focal volume $V_f \subset \Omega$:
    \begin{equation}
        \mathcal{J}_{\text{focus}}(\mathbf{x}) \;=\; \int_{V_f} \left\lVert \mathbf{E}(\mathbf{r};\mathbf{x},\lambda)\right\rVert^2 \, d\mathbf{r}.
    \end{equation}

    \item \textbf{Spectral sorting (color router).} For $K$ target wavelengths $\{\lambda_k\}_{k=1}^K$, we maximize the field intensity integrated over wavelength-specific target regions $\{R_k\}_{k=1}^K$:
    \begin{equation}
        \mathcal{J}_{\text{sort}}(\mathbf{x}) \;=\; \sum_{k=1}^{K} \int_{R_k} \left\lVert \mathbf{E}(\mathbf{r};\mathbf{x},\lambda_k)\right\rVert^2 \, d\mathbf{r}.
    \end{equation}

    \item \textbf{Mode conversion (waveguide mode converter).} We maximize the modal overlap between the output field on the cross-section $S_{\text{out}}$ and the desired target mode profile $\boldsymbol{\Psi}_{\text{tgt}}$.:
    \begin{equation}
        \mathcal{J}_{\text{mode}}(\mathbf{x}) \;=\;
        \left| \int_{S_{\text{out}}} \mathbf{E}(\mathbf{r};\mathbf{x},\lambda)\cdot \boldsymbol{\Psi}_{\text{tgt}}^{*}(\mathbf{r}) \, d\mathbf{r} \right|^2.
    \end{equation}
\end{itemize}

Although the objectives differ across tasks, they are all governed by the same frequency-domain Maxwell equation:
\begin{equation}
    \nabla \times \nabla \times \mathbf{E}(\mathbf{r}) - k_0^2 \epsilon(\mathbf{r}) \mathbf{E}(\mathbf{r})
    \;=\; -i\omega\mu_0 \mathbf{J}_{\mathrm{src}}(\mathbf{r}),
    \label{eq:maxwell}
\end{equation}
where $k_0 = 2\pi/\lambda$.

\subsection{The Adjoint-Solver Bottleneck}
To optimize the objectives defined above, gradient-based inverse design typically employs the \textit{adjoint method} (also known as the adjoint-variable method) to compute the sensitivity of the FoM with respect to the design parameters~\cite{Miller:EECS-2012-115, wang2018adjoint}. Let $\mathbf{g}=\nabla_{\mathbf{x}}\mathcal{J}$ denote the gradient. In the electromagnetic adjoint formulations, $\mathbf{g}$ can be written as a point-wise dot product between the forward field $\mathbf{E}{\mathrm{fwd}}$ and the adjoint field $\mathbf{E}{\mathrm{adj}}$:
\begin{equation}
\mathbf{g}(\mathbf{r}) \;\propto\; \mathrm{Re}\!\left[\mathbf{E}_{\mathrm{fwd}}(\mathbf{r}) \cdot \mathbf{E}_{\mathrm{adj}}(\mathbf{r})\right],
\label{eq:adjoint_grad}
\end{equation}
where the adjoint source term is determined by the particular FoM.

Importantly, independent of the chosen FoM, the dominant cost in adjoint-based optimization comes from repeatedly solving the full-wave Maxwell system in Eq.~(\ref{eq:maxwell}) to obtain the forward and adjoint fields. In 3D, high-resolution, and broadband settings, these repeated full-wave computations become the primary bottleneck, making conventional iterative optimization prohibitively expensive in practice.

\subsection{Neural Surrogate for Adjoint Gradients}
Our objective is to approximate the adjoint-to-gradient mapping $\mathcal{G}:\mathbf{x}\mapsto \nabla_{\mathbf{x}}\mathcal{J}$ using a unified neural surrogate $f_\theta$. Although the scalar FoM $\mathcal{J}$ differs across applications, the volumetric gradient field $\mathbf{g}=\nabla_{\mathbf{x}}\mathcal{J}$ encodes the local sensitivity information required by gradient-based optimizers. Accordingly, we formulate the problem as dense field prediction:
\begin{equation}
    \hat{\mathbf{g}} \;=\; f_\theta(\mathbf{x}) \;\approx\; \nabla_{\mathbf{x}}\mathcal{J}
    \;\in\; \mathbb{R}^{N_x \times N_y \times N_z}.
\end{equation}
In the remainder of the paper, we treat $f_\theta$ as a drop-in replacement for repeated adjoint solves within iterative optimization. The model architecture and training procedure, including supervision using gradients generated by full-wave solvers across the three objectives considered, are described in Sec.4 Methodology.

\section{Related Work}

\subsection{Adjoint Optimization in Photonic Inverse Design}
Adjoint-based optimization is a widely used approach for large-scale photonic and meta-optical inverse design, as it enables efficient gradient computation with only two full-wave simulations (forward and adjoint) per objective evaluation, largely independent of the number of design variables~\cite{Miller:EECS-2012-115, marzban2026inverse, dai2025shaping, pearson2025inverse, jokisch2025efficient, bi2025inversedesignedsiliconnitridenanophotonics}. In practice, however, the overall runtime remains dominated by repeatedly invoking full-wave simulations across many iterations and wavelengths, motivating efforts to reduce the cost of solver-in-the-loop optimization.

\subsection{Neural Operators and Surrogate Modeling}

Neural operators, including FNO, have emerged as mesh-agnostic surrogates for PDE-governed systems such as electromagnetics~\cite{azizzadenesheli2024neural, gu2022neurolight}. Most prior work has focused on surrogates that predict fields or spectra in order to reduce the cost of repeated simulations~\cite{seo2024wave, nam2026data, chen2022high, furat2025physics}. More recently, gradient learning has been investigated as an alternative, in which adjoint sensitivities required by gradient-based optimizers are directly regressed~\cite{jiang2019global, kang2025adjoint}. This approach is especially attractive because it predicts the gradient information used to compute iterative design updates.

\begin{figure*}[!htp]
    \centering
    \includegraphics[width=\linewidth]{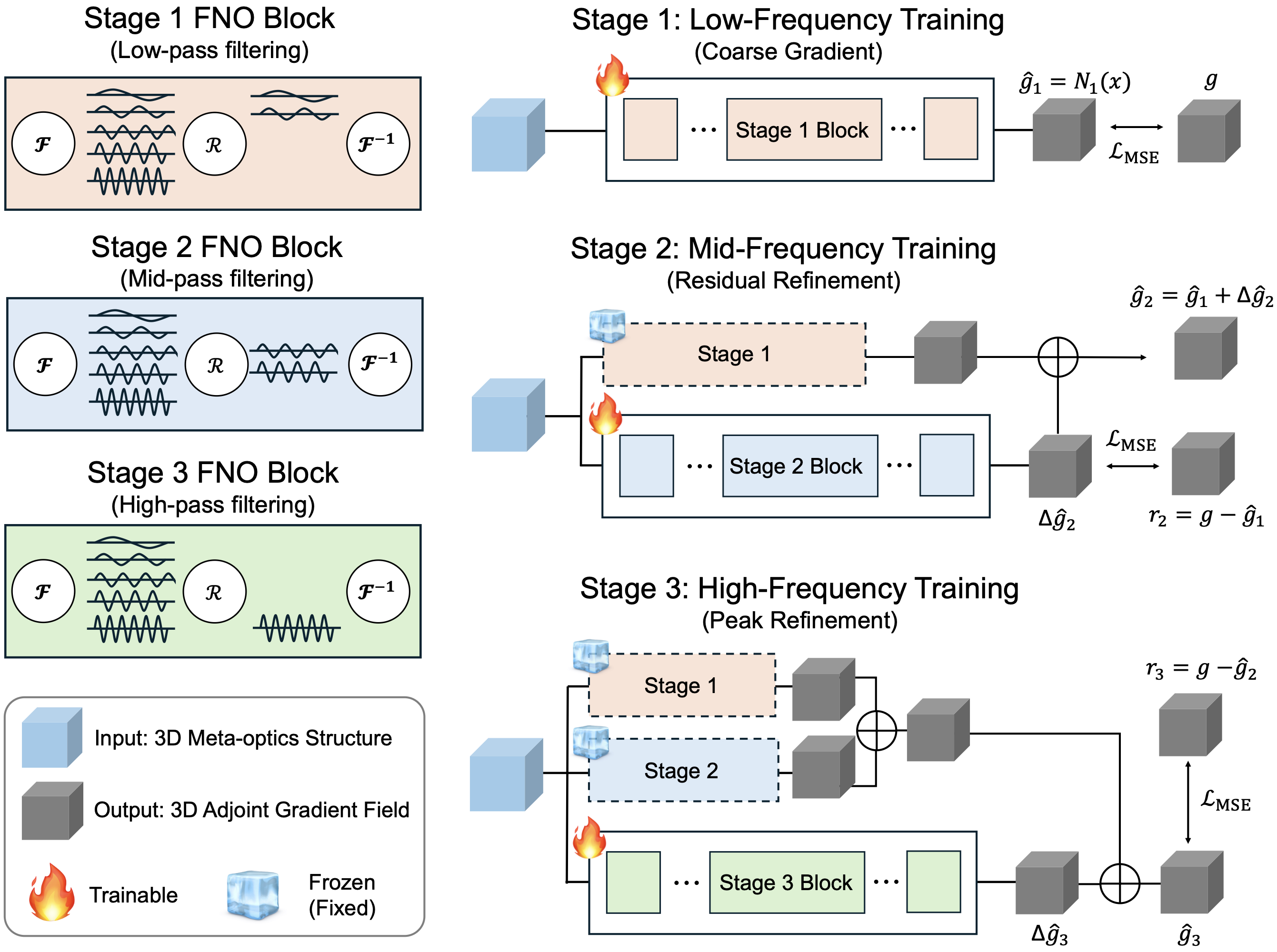}
    \caption{Stage-wise Fourier Neural Operator (SW-FNO).
    Overview of our stage-wise training scheme for predicting dense 3D adjoint gradient fields from a voxelized meta-optics structure. Each stage employs an FNO block operating in the Fourier domain, with frequency emphasis progressively shifted from low to mid to high bands (left: simplified frequency-selective FNO blocks). Training proceeds sequentially: Stage~1 learns a coarse, low-frequency prediction; subsequent stages are trained on residual targets to refine missing components, while earlier stages are frozen. The final prediction is obtained by summing the stage outputs, enabling accurate reconstruction across the full frequency spectrum while preserving sharp, high-frequency sensitivity peaks.}
\label{fig:fig2}
\end{figure*}

\subsection{Positioning and Distinctions of This Work}

Building on gradient learning with neural operators, we focus on \emph{3D volumetric} inverse design across diverse physics-based objectives, including focusing, spectral routing, and mode conversion. Extending neural-adjoint surrogates from low-dimensional or 2D problems to 3D introduces practical challenges related to representational capacity, preservation of high-spatial-frequency sensitivity, and scalability. We address these challenges through:
\begin{itemize}
\item \textbf{Volumetric setting and multi-objective coverage:} We develop a unified surrogate formulation that predicts 3D gradient fields on voxel grids and generalizes across multiple classes of objectives.
\item \textbf{Stage-wise refinement for high-frequency sensitivities:} We introduce a multi-stage refinement scheme, including spectral boosting, to better preserve sharp sensitivity peaks that are critical for faster convergence.
\item \textbf{Practical data construction for supervised gradient learning:} We design a controllable synthetic data-generation pipeline that produces diverse, smoothly varying topologies together with solver-generated gradients for supervised training.
\end{itemize}

\section{Methodology}

In this section, we present the Neural Adjoint Method for accelerating volumetric meta-optical inverse design. We introduce a \textit{Stage-wise Fourier Neural Operator} (SW-FNO) that predicts 3D adjoint gradient fields via coarse-to-fine residual refinement, improving the recovery of localized high-sensitivity structures. We then describe a solver-reduced optimization loop that updates the design using the inferred gradients, enabling rapid iterative refinement without repeatedly running full-wave forward/adjoint simulations.

\subsection{Stage-wise FNO (SW-FNO)}

To map a voxelized permittivity distribution $\mathbf{x} \in \mathbb{R}^{N_x \times N_y \times N_z}$ to a dense adjoint gradient field $\mathbf{g}$, the surrogate must represent electromagnetic phenomena including wave propagation, scattering, and resonance, the latter often producing sharp, high-spatial-frequency sensitivity peaks. In practice, we find that a single-stage FNO often underrepresents these high-frequency components, resulting in overly smooth gradient fields and, consequently, degraded optimization performance ~\cite{rahaman2019spectral}.

To address this limitation, we adopt a stage-wise refinement strategy that better matches the multi-scale nature of 3D volumetric scattering~\cite{qin2024toward}. The proposed SW-FNO performs progressive refinement: it first captures coarse, global trends and then successively adds residual corrections to recover fine-scale, high-spatial-frequency components~\cite{liu2024mitigating}. Formally,
\begin{equation}
\hat{\mathbf{g}} \;=\; \hat{\mathbf{g}}_{1} + \sum_{s=2}^{S} \Delta \hat{\mathbf{g}}_{s}, 
\qquad 
\hat{\mathbf{g}}_{1} = \mathcal{N}_{1}(\mathbf{x}),\;\;
\Delta \hat{\mathbf{g}}_{s}=\mathcal{N}_{s}\!\left(\mathbf{x}, \{\hat{\mathbf{g}}_{k}\}_{k=1}^{s-1}\right),
\label{eq:stagewise}
\end{equation}
where $S$ is the number of stages and $\Delta \hat{\mathbf{g}}_{s}$ denotes the residual correction predicted at stage $s$. This additive coarse-to-fine decomposition encourages later stages to focus on what earlier stages miss, enabling sharper recovery of high-frequency gradient features while preserving global consistency of the predicted field.

\subsubsection{Stage 1: Global Physics Learning}

The first stage uses a 3D FNO to capture the dominant low-spatial-frequency, global structure of the adjoint gradient field. These components reflect long-range interactions, wave propagation effects, and coarse sensitivity patterns throughout the volume. For the $l$-th Fourier layer, the feature update is defined as

\begin{equation}
\mathbf{v}_{l+1}(\mathbf{r}) =
\sigma\!\left(
W_l\,\mathbf{v}_l(\mathbf{r}) +
\mathcal{F}^{-1}\!\left(
R_l(\mathbf{k}) \odot \mathcal{F}(\mathbf{v}_l)(\mathbf{k})
\right)(\mathbf{r})
\right),
\label{eq:fno_layer}
\end{equation}

where $\mathcal{F}$ and $\mathcal{F}^{-1}$ denote the FFT and inverse FFT, $W_l$ is a pointwise linear projection, and $R_l$ is a complex-valued spectral weight tensor defined on the truncated set of Fourier modes (up to $k_{\mathrm{base}}$). Operating in the Fourier domain yields a global convolution that captures non-local, physics-driven correlations.

The Stage-1 predictor produces a coarse global estimate of the gradient field
\begin{equation}
\hat{\mathbf{g}}_{1} = \mathcal{N}_{1}(\mathbf{x}),
\label{eq:stage1_pred}
\end{equation}
where $\mathbf{x}$ denotes the voxelized device representation (e.g., permittivity) used as input to the neural operator. which provides a globally consistent initialization for subsequent residual stages and stabilizes training by avoiding the need to learn fine-scale corrections from scratch. Inevitably, with a limited mode budget ($k_{\mathrm{base}}$) and the spectral bias of operator networks, $\hat{\mathbf{g}}_{1}$ may smooth out boundary-aligned variations and localized resonant features that require higher spectral resolution.

\subsubsection{Stage 2: Mid-Frequency Residual Refinement}
\label{sec:stage2}

While Stage 1 captures dominant global sensitivity patterns, its prediction $\hat{\mathbf{g}}_{1}$ can under-represent mid-frequency structures around material interfaces, boundaries, and transition regions due to low-mode truncation and spectral bias. Stage 2, therefore, conditions on the Stage 1 estimate and predicts a residual correction, following the residual-learning principle ~\cite{he2016deep}:

\begin{equation}
\begin{aligned}
\mathbf{r}_{2} &= \mathbf{g} - \hat{\mathbf{g}}_{1}, \\
\Delta \hat{\mathbf{g}}_{2} &= \mathcal{N}_{2}\!\left([\mathbf{x}, \hat{\mathbf{g}}_{1}]\right), \\
\hat{\mathbf{g}}_{2} &= \hat{\mathbf{g}}_{1} + \Delta \hat{\mathbf{g}}_{2}.
\end{aligned}
\label{eq:stage2}
\end{equation}

where $\mathbf{g}$ denotes the ground-truth adjoint gradient field obtained from the FDTD solver. 

\subsubsection{Stage 3: High-Frequency Peak Refinement}
\label{sec:stage3}

Even after Stage~2 recovers boundary-aligned mid-frequency residuals, the target adjoint gradient field may still exhibit sharp, highly localized sensitivity peaks at higher spatial frequencies. These peaks often arise from abrupt spatial variations (e.g., resonance-induced confinement and interface discontinuities) and indicate fine-scale regions where topology updates must remain responsive.

Standard FNOs exhibit a low-frequency bias and, under finite mode truncation, tend to smooth spiky high-frequency components, which can underestimate gradient magnitudes near localized maxima. To better preserve these high-frequency extrema, Stage~3 predicts an additional residual correction conditioned on the Stage~2 estimate:
\begin{equation}
\begin{aligned}
\mathbf{r}_{3} &= \mathbf{g} - \hat{\mathbf{g}}_{2}, \\
\Delta \hat{\mathbf{g}}_{3} &= \mathcal{N}_{3}\!\left([\mathbf{x}, \hat{\mathbf{g}}_{2}]\right), \\
\hat{\mathbf{g}}_{\mathrm{final}} &= \hat{\mathbf{g}}_{2} + \Delta \hat{\mathbf{g}}_{3}.
\end{aligned}
\label{eq:stage3}
\end{equation}
By explicitly modeling the remaining residual, Stage~3 improves the recovery of high-magnitude localized peaks that are critical for fine-scale topology refinement, helping avoid attenuated sensitivities and preserving reliable update directions.

\subsection{Learning Adjoint Gradients from Full-Wave Simulations}
We train the surrogate $f_\theta$ on paired samples of voxelized device representations and their corresponding adjoint gradients, $\{(\mathbf{x}_i,\mathbf{g}_{i})\}$. Here, $\mathbf{g}_{i}$ is obtained from a conventional adjoint method using open-source FDTD simulations in Meep~\cite{OSKOOI2010687}: for each design $\mathbf{x}_i$, we run a forward simulation to evaluate the objective and an adjoint simulation to compute the volumetric sensitivity field. The learning problem is formulated as dense regression. Given $\mathbf{x}$, the model predicts the full 3D gradient field $\hat{\mathbf{g}} = f_\theta(\mathbf{x})$ and is optimized to match the solver-generated $\mathbf{g}$.

\section{Experiments}

\begin{figure*}[htp]
    \centering
    \includegraphics[width=\linewidth]{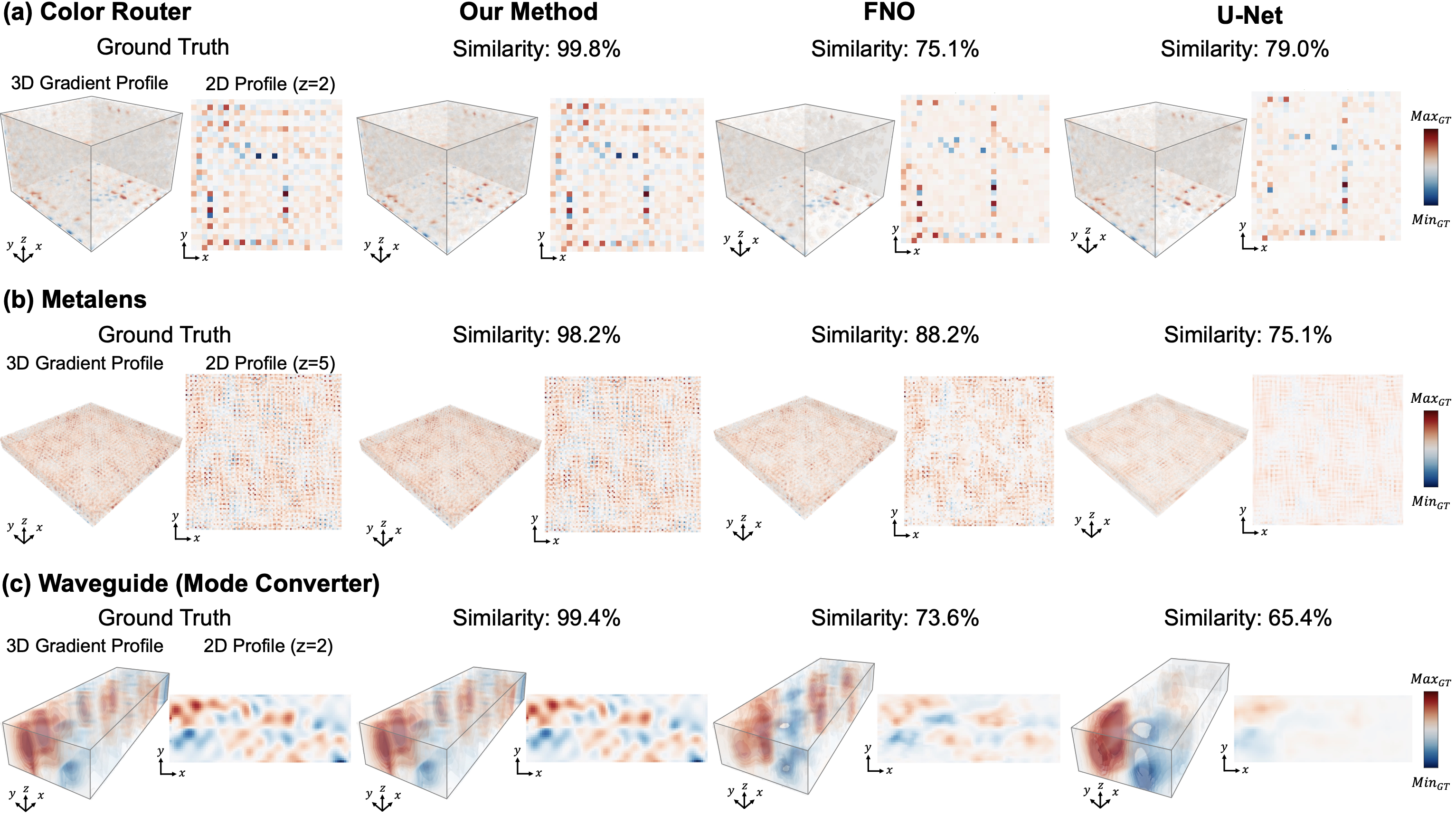}
\caption{\textbf{Qualitative comparison of volumetric adjoint-gradient predictions.} 
We visualize the ground-truth gradient fields alongside predictions from our method, standard FNO, and U-Net across three tasks: (a) RGB color router, (b) metalens, and (c) waveguide mode converter. 
Each panel displays a 3D volumetric rendering and a corresponding 2D cross-sectional slice at the indicated $z$-depth. 
While baseline models (FNO and U-Net) tend to produce over-smoothed fields that miss critical high-frequency components, our method accurately reconstructs sharp gradient peaks and fine-grained spatial features, achieving maximum cosine similarity scores exceeding 98\% across all benchmarks.}
\label{fig:fig4}
\end{figure*}

In this section, we empirically evaluate the Neural Adjoint Method on three volumetric meta-optical inverse-design tasks. Our experiments are designed to address the following questions:
\begin{itemize}
\item \textbf{Generalization (RQ1):} Can a single neural framework be applied to physically different inverse-design problems using a consistent architecture, without task-specific modifications?
\item \textbf{Efficiency (RQ2):} How much speedup does the proposed method provide relative to conventional FDTD-based adjoint optimization?
\item \textbf{Component Analysis (RQ3):} Do multi-stage refinement and high-frequency boosting improve gradient fidelity and final design performance, and what is the trade-off with model size and latency?

\end{itemize}

\subsection{Experimental Setup}
\subsubsection{Datasets}

We construct a domain-specific dataset consisting of paired permittivity volumes and adjoint sensitivity fields. Unlike generic PDE benchmarks that emphasize smooth state solutions, our supervision uses dense adjoint gradient fields as labels, which exhibit sharp, localized sensitivity features induced by interference and resonant effects in meta-optics structures.

Depending on the task and labeling cost, we generate 600 waveguide samples, 1{,}000 metalens samples, and 5{,}000 color-router volumetric samples. For each sample, we compute the ground-truth adjoint gradient field using open-source FDTD Meep via paired forward and adjoint simulations, and form voxel-wise sensitivities~\cite{OSKOOI2010687}. We use fixed train/validation/test splits with a fixed random seed for reproducibility. 

\subsubsection{Data Processing and Analysis}
All samples are represented on a fixed voxel grid and normalized to a consistent permittivity range. Gradient labels are stored as dense 3D tensors aligned with the design region, and voxels outside the design domain are masked consistently across tasks.

For evaluation, we report (i) \textit{functional fidelity} as the ratio of the final FoM achieved by our method to the FDTD benchmark, and (ii) \textit{structural fidelity} as the cosine similarity between the final voxelized designs.

\subsubsection{Baselines}
We evaluate the Neural Adjoint Method against the following baselines, covering both physics-grounded optimization and learning-based surrogates. 

\begin{enumerate}
    \item \textit{Adjoint Method (FDTD, Reference).} A conventional adjoint optimization pipeline that computes gradients via paired forward/adjoint full-wave simulations and updates the design with a gradient-based optimizer. This serves as a physically accurate reference for optimization and provides the benchmark FoM for comparison.

    \item \textit{One-Stage FNO.} A vanilla FNO that predicts the full 3D adjoint gradient field in a single pass, without any stage-wise refinement. 
    
    \item \textit{Triple-Stage FNO (SW-FNO).} Our proposed stage-wise FNO that cascades three independent FNO blocks with progressively larger Fourier-mode budgets, enabling a coarse-to-fine refinement from low- to high-frequency content. 
    
    \item \textit{U-Net.} 3D U-Net with an encoder–decoder architecture and skip connections, trained to predict the dense adjoint gradient field. 
    
\end{enumerate}


\subsection{Adjoint Gradient Inference on Diverse Benchmarks (RQ1)}

\begin{table}[h]
\centering
\caption{Comparison of gradient prediction performance (Cosine Similarity).}
\label{tab:summary_results}
\resizebox{\columnwidth}{!}{%
\begin{tabular}{l|cc|cc|cc}
\toprule
\multirow{2}{*}{\textbf{Model}} & \multicolumn{2}{c|}{\textbf{Color Router}} & \multicolumn{2}{c|}{\textbf{Metalens}} & \multicolumn{2}{c}{\textbf{Waveguide}} \\ \cmidrule(lr){2-3} \cmidrule(lr){4-5} \cmidrule(lr){6-7} 
 & \textbf{Mean (Std)} & \textbf{Max} & \textbf{Mean (Std)} & \textbf{Max} & \textbf{Mean (Std)} & \textbf{Max} \\ \midrule
U-Net & 61.4 (13.7) & 86.4 & 53.6 (6.5) & 69.1 & 32.6 (23.9) & 71.9 \\
FNO & 66.3 (13.1) & 89.7 & 73.5 (10.3) & 90.0 & 46.4 (25.2) & 80.5 \\
\textbf{Ours} & \textbf{82.3(13.7)} & \textbf{99.8} & \textbf{87.6 (7.5)} & \textbf{98.2} & \textbf{87.9 (30.3)} & \textbf{99.4} \\ \bottomrule
\end{tabular}%
}
\end{table}

We evaluate the proposed framework on three representative high-dimensional inverse design problems: an RGB color router, an achromatic metalens, and a waveguide mode converter. To rigorously quantify the capability of the proposed neural adjoint method, we analyze both structural fidelity and functional performance.

To quantify gradient fidelity, we compute the cosine similarity between the surrogate-predicted gradient fields and the rigorous FDTD adjoint gradients, which directly measures the alignment of update directions used by gradient-based optimizers. 

As summarized in Table~\ref{tab:summary_results}, our method consistently achieves the highest cosine similarity across all three tasks, outperforming both standard one-stage FNO and 3D U-Net baselines. Qualitatively, Fig.~\ref{fig:fig4} further supports this conclusion. Baseline models tend to produce over-smoothed gradient fields that attenuate localized extrema, whereas our method reconstructs sharper peak structures and fine-grained spatial features across all benchmarks.

\subsection{Optimization on Diverse Benchmarks (RQ1 \& RQ2)}
We evaluate the proposed Neural Adjoint Method on three representative volumetric inverse-design tasks: an RGB color router, an achromatic metalens, and a waveguide mode converter. These benchmarks span distinct objectives providing a stringent test of whether one neural framework can be applied across heterogeneous design problems without task-specific architectural modifications.

\begin{figure}[!htp]
    \centering
    \includegraphics[width=\linewidth]{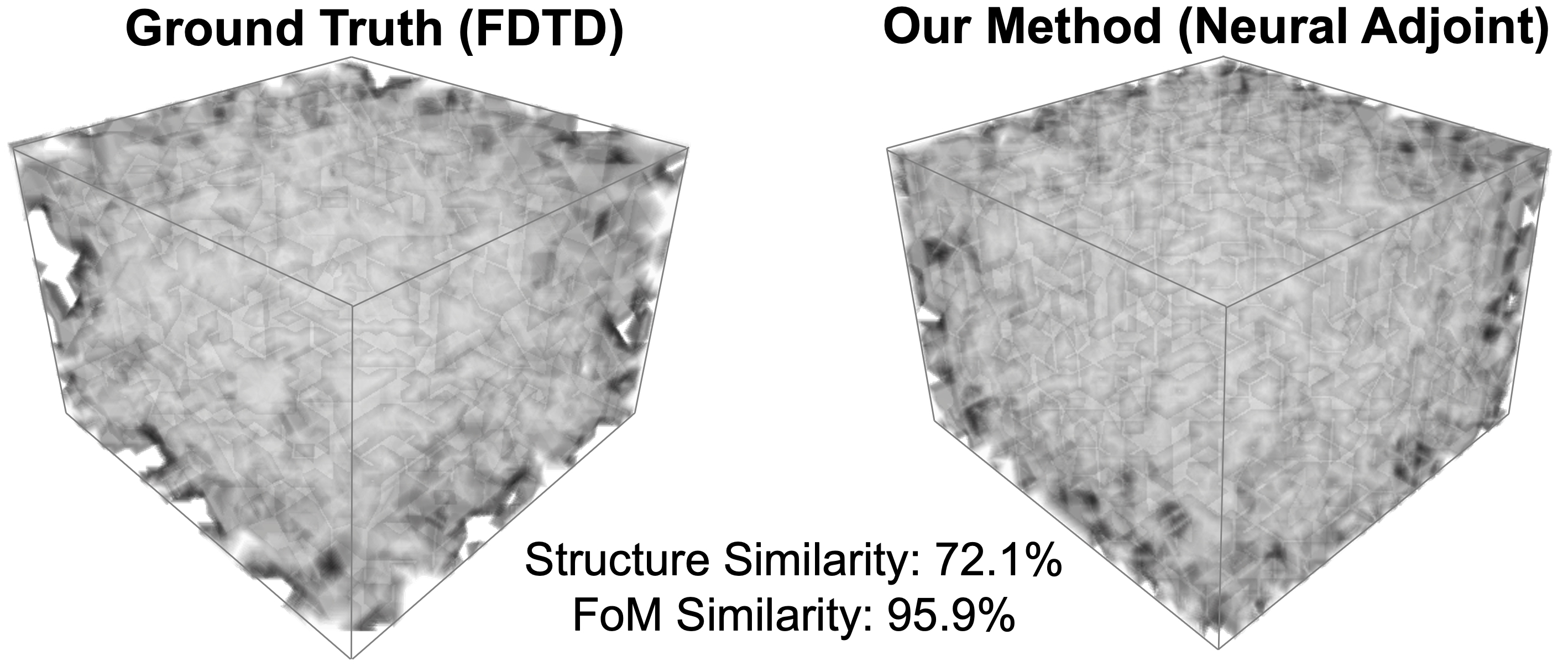}
    \caption{Qualitative comparison of final color-router designs. Volumetric renderings of the optimized permittivity distribution obtained by the FDTD-based adjoint optimization (left, ground truth) and by our Neural Adjoint method (right). Although the resulting geometries are not necessarily the same (structure similarity: 72.1\%), our method achieves comparable functional performance under the same objective (FoM similarity: 95.9\%).}
    \label{fig:fig5}
\end{figure}

To assess end-to-end utility for inverse design, we report two complementary outcomes: (i) the final design performance achieved by surrogate-driven optimization and (ii) the total wall-clock design time required to reach the final design. For performance, we compare the FoM obtained by our solver-free optimization against a  FDTD adjoint optimization. We report Relative Optimality (FoM, structure similarity) as the ratio of the achieved efficiency to the FDTD baseline, and summarize results across tasks in Table~\ref{tab:main_results_halfcol}. Qualitative examples of optimized color router and their corresponding responses are shown in Fig.~\ref{fig:fig5}.

\begin{table}[!ht]
    \centering
    \caption{Relative Optimality and Design Speed. We compare our neural adjoint-optimized designs against the FDTD benchmark. All methods are run for a fixed number of optimization iterations: 50 (Color Router), 100 (Metalens), and 30 (Waveguide).}
    \label{tab:main_results_halfcol}

    \setlength{\tabcolsep}{5pt}
    \renewcommand{\arraystretch}{1.15}

    \resizebox{\columnwidth}{!}{%
    \begin{tabular}{l cc cc c}
        \toprule
        \multirow{2}{*}{\textbf{Task}} &
        \multicolumn{2}{c}{\textbf{Rel. Opt. (\%)}} &
        \multicolumn{2}{c}{\textbf{Design Time}} &
        \multirow{2}{*}{\textbf{Speedup}} \\
        \cmidrule(lr){2-3} \cmidrule(lr){4-5}
         & \textbf{FoM} & \textbf{Struct.} & \textbf{FDTD} & \textbf{Ours} & \\
        \midrule
        \textbf{Color Router} & $\mathbf{95.9}$ & $72.1$ & \textit{3.4 h} & \textit{1.0 s} & $\times 12{,}194$ \\
        \textbf{Metalens} & $\mathbf{86.0}$ & $83.6$ & \textit{40.1 h} & \textit{1.8 s} & $\times 80{,}200$ \\
        \textbf{Waveguide} & $\mathbf{98.8}$ & $91.2$ & \textit{11.5 h} & \textit{0.7 s} & $\times 59{,}382$ \\
        \bottomrule
    \end{tabular}%
    }
\end{table}

Across all three benchmarks, our method achieves high relative optimality while substantially reducing design time. The resulting designs remain close to the FDTD-optimized solutions in terms of final performance, despite eliminating repeated full-wave forward/adjoint solves during optimization. This indicates that the neural adjoint surrogate supports a shared optimization pipeline that transfers across disparate meta-optical objectives.

\begin{table*}[htp]
\centering
\caption{Ablation of the number of refinement stages in SW-FNO (1--4). We report the stage-wise operating modes together with parameter count, inference latency, and final optimized efficiency.}
\label{tab:stage_modes_ablation}
\setlength{\tabcolsep}{6pt}
\begin{tabular}{cccccc}
\hline
\textbf{\#Stages} & \textbf{Stage-wise Modes} & \textbf{\#Params (M)} $\downarrow$ & \textbf{Latency (ms)} $\downarrow$ & \textbf{Eff. (\%)} $\uparrow$ \\
\hline
1 &
S1: \texttt{Low} &
$7.09$ & $4.8$ & $73.5$ \\

2 &
S1: \texttt{Low} $\rightarrow$ S2: \texttt{High} &
$14.71$ & $7.8$ & $73.1$ \\

3 (Ours) &
S1: \texttt{Low} $\rightarrow$ S2: \texttt{Mid} $\rightarrow$ S3: \texttt{High} &
$21.80$ & $10.9$ & $87.6$ \\

4 &
S1: \texttt{Low} $\rightarrow$ S2: \texttt{Low-Mid} $\rightarrow$ S3: \texttt{High-Mid} $\rightarrow$ S4: \texttt{High} &
$42.53$ & $14.0$ & $88.7$ \\
\hline
\end{tabular}
\end{table*}

\subsection{Analysis of Computational Efficiency (RQ2)}
The primary advantage of our method is the elimination of the simulation bottleneck.
Compared to conventional solver-in-the-loop adjoint optimization, our approach yields orders-of-magnitude speedups on all tasks (Table~\ref{tab:main_results_halfcol}), reducing design time from hours to seconds. While a single iteration of 3D FDTD simulation for color router takes approximately minutes to hours on a workstation (CPU: AMD 9995WX), the Neural Adjoint inference takes only milliseconds on a single GPU (NVIDIA H200).

\begin{figure}[htp]
    \centering
    \includegraphics[width=0.8\linewidth]{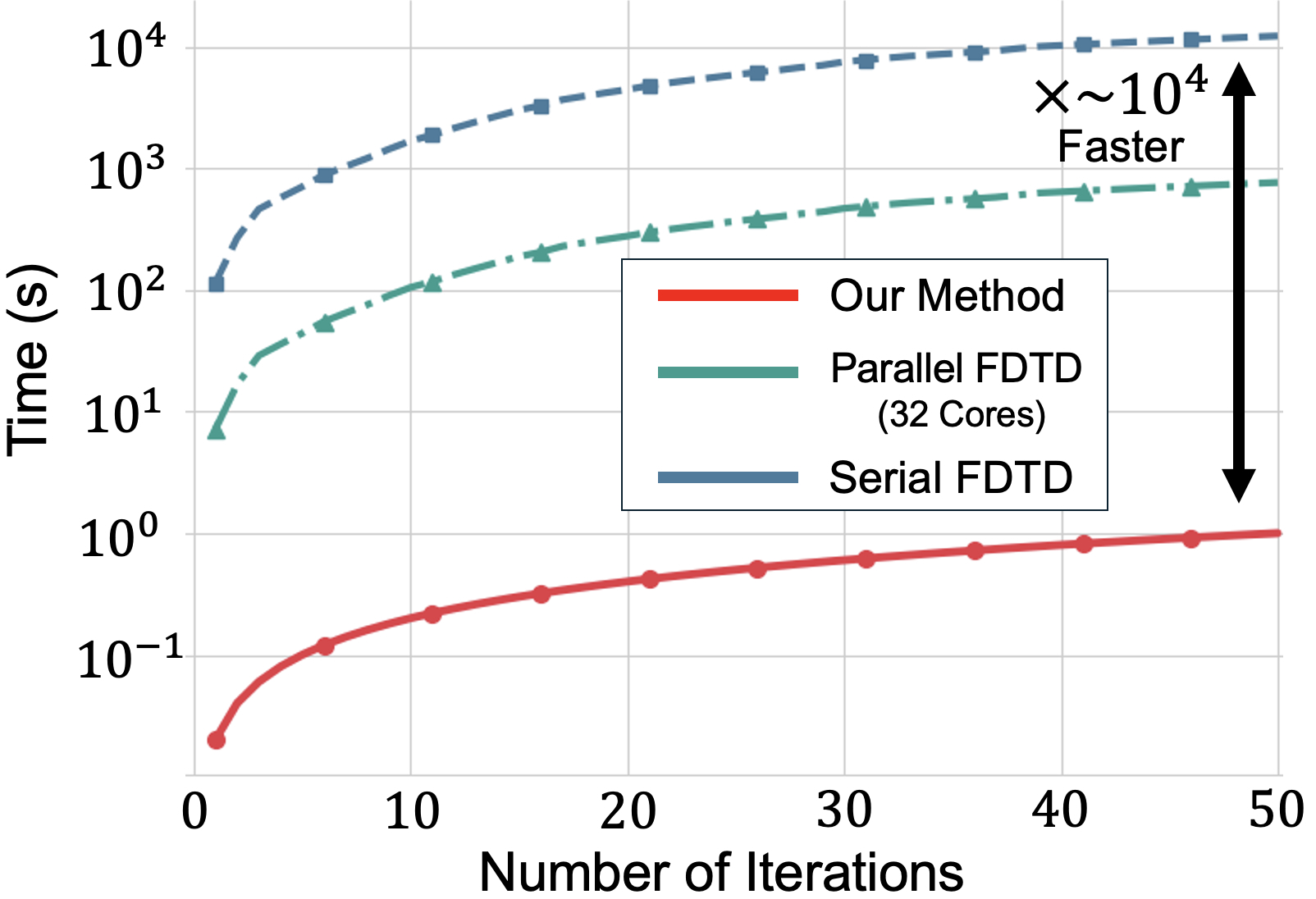}
    \caption{Optimization time comparison. Wall-clock time versus the number of optimization iterations for our Neural Adjoint method and FDTD-based adjoint optimization. Across iteration, our method remains orders of magnitude faster, achieving up to $\sim 10^{4}\times$ speedup over serial FDTD.}    
    \label{fig:fig6}
\end{figure}

\subsection{Effect of Multi-Stage Refinement and Spectral Boosting (RQ3)}
Table~\ref{tab:stage_modes_ablation} shows that increasing the number of refinement stages improves the final efficiency, with the largest gain achieved by three stages and only marginal improvement beyond that. This trend is consistent with the spectral bias of single-pass operator networks, which tend to under-represent sharp, high-spatial-frequency sensitivity peaks that are important for effective topology updates. By introducing low-to-high (coarse-to-fine) residual refinement with progressively higher-frequency emphasis, the multi-stage design better restores these localized extrema, reducing gradient-field error and yielding higher-performing optimized devices.

Notably, moving from three to four stages increases efficiency by only $1.1\%$, while substantially increasing both the number of parameters and inference latency. Given this trade-off, we adopt the three-stage configuration as our default, which provides a favorable balance between accuracy and computational overhead.

\section{Limitations and Ethical Considerations}
\subsection{Limitations}

Our training data are generated under a fixed simulation configuration and from a controlled distribution of synthetic structures. As a result, the learned surrogate necessarily reflects the assumptions and coverage of this training domain and may not directly generalize to substantially different geometries, material models, fabrication constraints, or experimental conditions without additional data and/or domain adaptation. Moreover, although neural adjoint inference can substantially reduce reliance on full-wave simulations during iterative design, the current study is not yet fully solver-free in the broader sense, and our evaluation is limited to relatively compact volumetric domains.

A direction for future work is to explicitly leverage the scalability of neural operators to enable cross-scale generalization: training on smaller, simulation-feasible subdomains and transferring the learned operator to predict gradients on much larger volumes. If successful, such a strategy could further decouple optimization throughput from the cost of full-wave simulation and enable practical, fully solver-free volumetric optimization pipelines. 

This work is a small step toward replacing expensive full-wave adjoint loops with learned gradient inference, but it marks meaningful progress toward real-time volumetric meta-optics inverse design.

\subsection{Ethical Considerations}
This research uses only synthetic data generated by full-wave electromagnetic simulations and does not involve human participants, personally identifiable information, or user-generated content; therefore, it does not raise privacy or informed-consent concerns. The proposed technology is intended to accelerate the design of optical components for display, imaging, and communication systems. While advanced optics could in principle be integrated into surveillance or military hardware, the method itself is a general-purpose optimization and surrogate-modeling tool and does not inherently enable misuse beyond general engineering capabilities.

\section{Conclusion}
We introduced the \textit{Neural Adjoint Method}, a neural-operator-based approach that targets a central bottleneck in computational meta-optics: the repeated full-wave forward/adjoint simulations required for 3D adjoint optimization. Instead of predicting only scalar objectives, our method learns to directly infer dense 3D adjoint gradient fields on voxelized device domains, enabling rapid iterative refinement while preserving the physically meaningful update directions that drive topology optimization.

We proposed a Stage-wise Fourier Neural Operator that performs coarse-to-fine residual refinement with progressively higher-frequency emphasis, improving recovery of localized, high-magnitude sensitivity peaks that strongly influence topology updates. 

To support rigorous analysis and reproducibility, we construct a domain-specific dataset from a conventional adjoint method using full-wave FDTD simulations, pairing voxelized designs with solver-generated volumetric adjoint gradients. Across three physically distinct inverse-design benchmarks (RGB color router, achromatic metalens, waveguide mode converter), our surrogate-driven optimization reaches solutions that remain close to FDTD adjoint optimization in final FoM, while reducing design time from hours to seconds. 
These results suggest a practical route toward interactive, general-purpose volumetric meta-optics design, bridging advances in neural operators with the stringent fidelity requirements of electromagnetic inverse design.

\newpage

\section*{Acknowledgments}
This work was supported by the National Research Foundation of Korea (NRF) grant funded by the Korean government (MSIT) (RS-2024-00338048, RS-2024-00414119, RS-2025-25463760, RS-2026-25487796); by the Ministry of Science and ICT (MSIT) of the Republic of Korea under the Global Research Support Program in the Digital Field (RS-2024-00412644) supervised by the Institute of Information and Communications Technology Planning \& Evaluation (IITP); by the Culture, Sports and Tourism R\&D Program through the Korea Creative Content Agency grant funded by the Ministry of Culture, Sports and Tourism in 2024 (RS-2024-00332210); by the Artificial Intelligence Graduate School Program (No. RS-2020-II201373, Hanyang University) supervised by IITP; by the artificial intelligence semiconductor support program to nurture the best talents (IITP-(2025)-RS-2023-00253914) funded by the Korean government; and by MSIT (RS-2025-02218723, RS-2025-02283217).

\bibliographystyle{ACM-Reference-Format}
\bibliography{sample-base}

\end{document}